\documentclass[conference]{IEEEtran}
\IEEEoverridecommandlockouts
\usepackage{amsmath,amssymb,amsfonts}
\usepackage{textcomp}
\usepackage{xcolor}
\usepackage{microtype}
\usepackage{multicol}
\usepackage{graphicx}
\usepackage{algorithm} 
\usepackage{algpseudocode} 
\usepackage[numbers,sort&compress]{natbib}
\usepackage{hyperref}
\usepackage{times}
\usepackage{latexsym}
\usepackage{pdfpages}
\usepackage{booktabs}
\usepackage{placeins}
\usepackage{adjustbox}

\usepackage{soul}
\def\BibTeX{{\rm B\kern-.05em{\sc i\kern-.025em b}\kern-.08em
    T\kern-.1667em\lower.7ex\hbox{E}\kern-.125emX}}
\begin{document}

\title{Improving Imbalanced Text Classification with Dynamic Curriculum Learning
}


\author{\IEEEauthorblockN{Xulong Zhang, Jianzong Wang$^\ast$\thanks{$^\ast$Corresponding author: Jianzong Wang, jzwang@188.com.}, Ning Cheng, Jing Xiao}
\IEEEauthorblockA{\textit{Ping An Technology (Shenzhen) Co., Ltd.}}
}

\maketitle

\begin{abstract}
    Recent advances in pre-trained language models have improved the performance for text classification tasks. However, little attention is paid to the priority scheduling strategy on the samples during training. Humans acquire knowledge gradually from easy to complex concepts, and the difficulty of the same material can also vary significantly in different learning stages. Inspired by this insights, we proposed a novel self-paced dynamic curriculum learning (SPDCL) method for imbalanced text classification, which evaluates the sample difficulty by both linguistic character and model capacity. Meanwhile, rather than using static curriculum learning as in the existing research, our SPDCL can reorder and resample training data by difficulty criterion with an adaptive from easy to hard pace. The extensive experiments on several classification tasks show the effectiveness of SPDCL strategy, especially for the imbalanced dataset. 
\end{abstract}


\section{Introduction}
	
Recent attention has been devoted to various training tasks for large neural networks pre-training, deep model architectures and model compression \cite{xu2021beyond,tang2022avqvc,lin2020birds,wang2020integrating,Zhang2021Cyclegean}. 
Imbalanced training data is common in real applications, which poses challenges for text classifiers. In this work, we study how to leverage curriculum learning (CL) to tackle the imbalanced text classification.
	
	
 CL \cite{bengio2009curriculum}, which mimics the human learning process from easy to hard concepts, can improve the generalization ability and convergence rate of deep neural networks. However, such strategies have been largely ignored for imbalanced text classification tasks due to the fact that difficulty metrics like entropy are static and fixed. So these metrics ignore that the difficulty of data samples always vary while training. Therefore the sample importance weights should be adaptive rather than fixed. 
To address this challenge, we proposed the SPDCL framework, which utilizes both the nuclear-norm and model capacity to evaluate the difficulty of training samples. Nuclear-norm, the sum of singular values, is used to constrain the low state of the matrix. From linear algebra, a matrix is low-norm and contains a large amount of data information, which can be used to restore data and extract features. From our experiments, the nuclear norm of the training sample is a process of first decreasing and then increasing until stable fluctuation from pre-training to fine-tuning stage.
	
That is to say, for the sample, it is the process of firstly removing the noise and then learning the deeper semantic features of the specific scene data. For samples whose noise points are far from the normal value or whose feature points are well distinguished, the nuclear norm changes more drastically, on the contrary, the nuclear-norm changes more smoothly. Specifically, simple examples are easier to recognize along with slight changes in sentence features during different training time, the nuclear-norm of which change more drastically. For different training epochs, we dynamically adopt their corresponding nuclear-norm production to calculate a difficulty score for each example in the train set. In each training phase, we will calculate the change in the kernel norm of the samples based on the relative position at the current moment, and update the samples as a new difficulty level. After reordering the samples in order from simple to complex in each round, we resampled the data to achieve the goal of learning simple samples first and then gradually increasing the difficulty of learning.
	
 Our contribution can be summarized as follows: 1) We proposed an adaptive sentence difficulty criterion, consisting of both linguistically motivated features and learning and task-dependent features. 
		2)We proposed a novel dynamic CL strategy that consists of re-calculating the difficulty criterion and resampling each epoch. 
		3)We prove the effectiveness of SPDCL in the context of fine-tuning language model for a series of text classification tasks including multi classification, multi label classification and text matching. 

	\section{Methodology}
	
	To elaborate our methods, we utilize the BERT-Base model \cite{devlin2018bert} as our example. The same method is also compatible to many other encoders, such as RoBERTa \cite{2019RoBERTa}, etc. Following BERT, the output vector of the $[\text{CLS}]$ token can be used for classification and other tasks since it can represent the kernel feature of all the sentence. So, we just add a linear layer on the $[\text{CLS}]$ sentence representation, then fine-tune the entire model with our CL strategy. But before each iteration, we would calculate the difficulty metric on the whole tokens along with each epoch rather than only on $[\text{CLS}]$ token to measure the difficulty score of each input sample.
	
	Formally, when we input the text sequence $\{[\text{CLS}], x_{1}, \ldots, x_{m}, [\text{SEP}]\} $ or $\{[\text{CLS}], 
	x_{11}, \ldots, x_{1m}, [\text{SEP}], x_{21}, \ldots, x_{2n}, [\text{SEP}]\}$ into the model, we can get the output token matrix $\{h_{\text{CLS}}, h_{x_{1}}, \ldots, h_{x_{m}}, h_{\text{CLS}}\}$ or $\{h_{\text{CLS}}, h_{x_{11}},\ldots, h_{x_{1m}},h_{\text{SEP}}, h_{x_{21}}, \ldots, h_{x_{2n}}, h_{\text{SEP}}\}$. Then, we compute the nuclear-norm for the whole token-level representations for each sample per epoch while training that represents the present information capacity. Before fine-tuning, the text representation obtained from the pre-trained model contains some redundant information, especially in some specific domain datasets. In the fine-tuning phase, the model further gets better text representation for specific tasks, specifically a process of removing redundant features to extract critical information and then further learning more finely granular features. In fact, in our experiments, we also find the phenomenon that nuclear-norm of samples firstly decreased and then gradually increased. Simple samples vary more dramatically, and complex samples are more difficult to learn about fine-grained and high-dimensional characteristics. We calculate the difference between the current nuclear-norm and the previous nuclear-norm for each sample, and sorted with an easy-to-difficult fashion per epoch with our curriculum learning strategy. 
	
	Finally, we regard the final layer embedding of first token, $[\text{CLS}]$, as the representation of the whole input sequence. On the one hand, in sequence classification and text matching tasks, we put the last layer matrix into a softmax activation function to predict its category label. On the other hand, in multi label classification tasks, we take a general approach to verifying the generality of the CL strategy.
	
	\subsection{Self-paced Dynamic Curriculum Learning}
\begin{figure}[!ht]
		\center
		\centerline{\includegraphics[width=\linewidth]{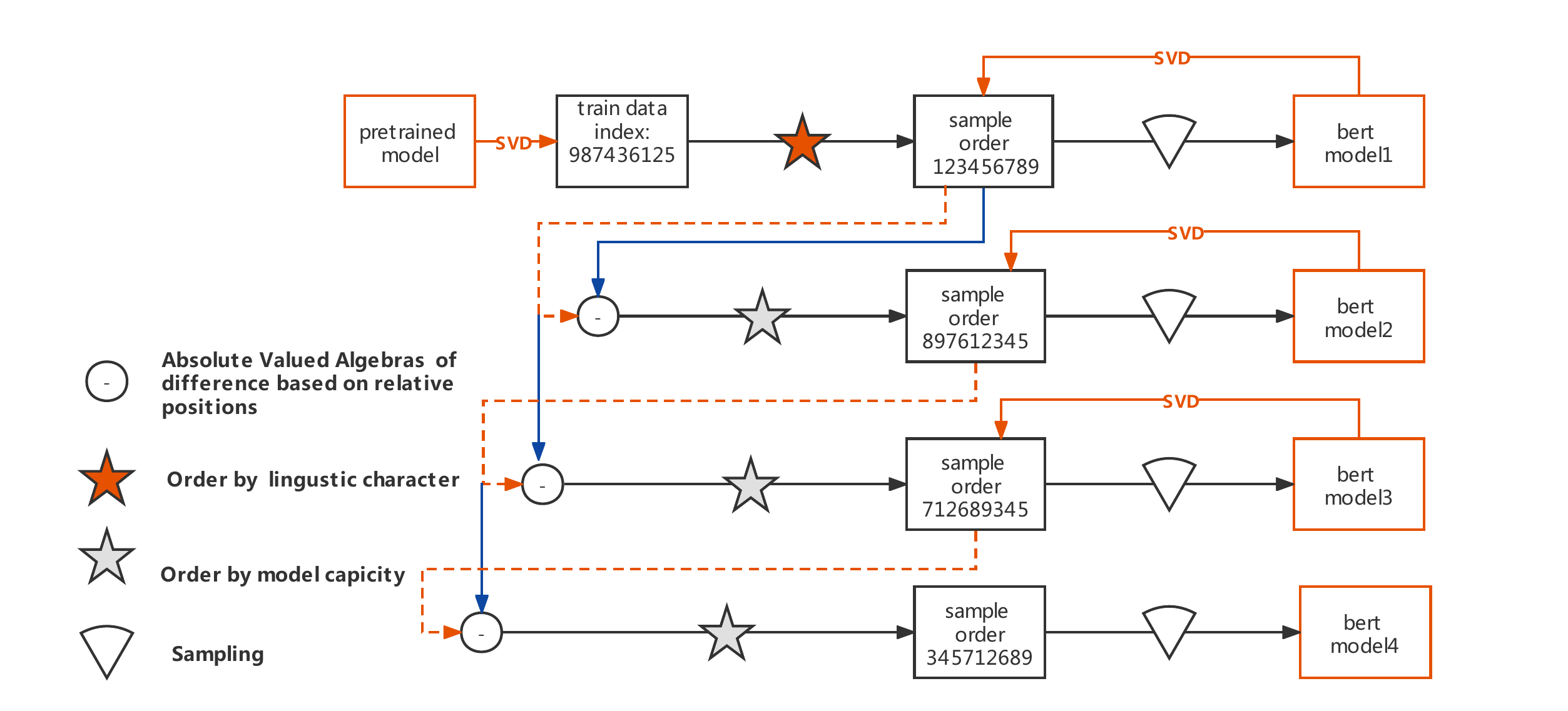}}
		\caption{Structure of SPDCL.It contains two key modules, difficulty criterion and curriculum arrangement. In our method, we divide the difficulty criterion into two modules, linguistic difficulty criterion before training and difficulty criterion based on model capacity while training. }
		\label{fig:structure}
	\end{figure}

	In this section, we would introduce our training strategy, Self-Paced Dynamic Curriculum Learning (SPDCL),the process of which is shown in Figure \ref{fig:structure}. More details about the pseudo-code can be seen in Algorithm 1. In general, curriculum learning contains two key modules, difficulty criterion and curriculum arrangement. In our method, we divide the difficulty criterion into two modules, linguistic difficulty criterion before training and difficulty criterion based on model capacity while training. Since we think the curriculum arrangement should change along with difficulty criterion. At first, difficulty criterion relies more on linguistic character when the model has not seen the data of downstream task and the feedback of the current model is not reliable. Later, when going into the formal training, the current state of the model itself should plays a more important role for difficulty criterion. What's more, there are always associations between difficulty criteria that are not independent in every time period.
	
\subsection{Linguistic difficulty criterion}
	
\begin{table}
	\centering
	\fontsize{8}{13}
	\selectfont
	\caption{The relationship between text length and nuclear-norm of sentence of train data on the AAPD dataset obtained from pre-trained model before fine-tuned procedure. Sentence bins are sorted by sentence difficulty based on linguistic character.It is obvious that longer sentences that containing more lexical-semantic and syntactic information tend to have larger nuclear-norm value before fine-tuning process.}	
	\begin{tabular}{c|cccccc}
		\hline
		Bins & 1&2&3&4&5&6\\
		\hline
		 Avg text length&440&702&881&1052&1260&1594\\
		\hline
	\end{tabular}
	\label{table:criterion}
\end{table}
	
	 The difficulty of sample sentence shows that it can be divided into two parts: heuristics metrics like sentence length, word frequency, and model based metrics of the specific task, which can be loss, accuracy, precision and F1 score, etc. \cite{wang2022drvc,platanios2019competence}. However, these methods divide linguistic character and model based character of a textual sample that both vital to a textual sample in different stages. For example, in the early stage, the teacher can rely more on linguistic character. According to the linguistic character, most NLP systems have been taking advantage of distributed word embeddings to capture the syntactic features of a word both in the direction and the norm of a word vector. While, the norm of the sentence matrix is rarely considered and explored in the computation of the difficulty metric for a textual sample. In contrast, the traditional word based difficulty metric does not have a particularly good way to directly obtain the difficulty metric of sentence-level representations, which generally average or sum all word norm, but this is not equivalent to the metric of representation of the modeled sentences displayed. Taking into account the polysemy and location of the word in the sentence and the deeper syntactic structure, semantic relationships, we proposed nuclear-norm based sentence difficulty criterion.
	
	The nuclear-norm is the sum of the singular values of the matrix, a classic representation of the amount of information that plays a vital role in natural language processing tasks\cite{liu2020norm}.In our previous experiment, we also find consistent phenomenon that the nuclear-norm of the sentence is related to the length of sentence shown in Table \ref{table:criterion}. Specially, longer sentences tend to have larger nuclear-norm values. At the initial learning stage when the model weights obtained from pre-trained model that has not yet seen the train dataset for downstream tasks, linguistic characters of a textual sample itself matters how hard it is for the downstream task. In our opinion, it is easier to train and predict the sentences with smaller nuclear-norm, as those sentences consist less information and easier syntactic structures. Compared with those sentences with smaller nuclear-norm, those harder sentences always with larger nuclear-norm that require being able to recognize the semantics of their component parts, it is also necessary to identify the syntactic structure and semantic relations between these component parts. To illustrate, we extracted and translated two sentences from sentiment classification task of \cite{li2016emotion}, Easier sentence 1: \textit{Today and everyone party is very happy}. Harder sentence 2: \textit{It's a pleasure to party with everyone today, but it's even more lonely and sad after the break. }
	
	Moreover, nuclear-norm value is always to constrain the low rank of the matrix which contains a lot of redundant information for sparse data that can be used to recover data and extract characteristics. \cite{mu2017representing} proposes that represent sentences as low-rank subspaces is efficient for downstream task, which is also consistent with our experiment result that the nuclear-norm of sentence representation after fine-tune stage is smaller than gotten from pre-trained stage. 
	
	Compared to previous studies in word norm based difficulty metric and other sentence norm like L2-norm and L1-norm, nuclear-norm mines and explores the information of error sentences. Word-based norm \cite{liu2020norm} and L2-norm and L1-norm both are vector-based element-wise norm. They are always assumed that words in a sentence are independent. That means, the results of those norm of a vector are not changed when we randomly exchange the location of the elements of the word vector while the semantic information of the whole sentence may have changed dramatically. Therefore, they cannot well exploit the correlation between words in words of a sentence which is important for text representation and classification. However, the nuclear-norm represents the values of a matrix. The operation is intended for coping with the information on the entire error sentence matrix rather than handling single word independently. If the position of the elements is exchanged, the value of the nuclear-norm of the matrix will change. Therefore, it can efficiently exploit the spatial structure information.
	
	So we first regard the nuclear norm of sentence representation matrix obtained from the pre-trained model as an initial difficulty score.

	\begin{algorithm}
		\caption{Self-Paced Dynamic Curriculum Learning (SPDCL)} 
		
		\textbf{Require: Data Corpus}$\ D=\left\{ \left\langle \mathbf{x}^n, \mathbf{y}^n \right\rangle \right\}_{n=1}^{N};$ Classification system $\theta;$ Bin numbers $k$
		\begin{algorithmic}[1]
			\State Compute Nuclear Norm as $\left\{d_1\left(\mathbf{x}^n \right)\right\}_{n=1}^N$ on pertained model
			\State Sort $\left\{\mathbf{x}_{1}^{n}\right\}_{n=1}^{N}$  by $\left\{d_1\left(\mathbf{x}^{n}\right)\right\}_{n=1}^{N}$ from small to large
			\State Divide $\left\{\mathbf{x}_{1}^{n}\right\}_{n=1}^{N}$ into $k$ parts and re-sample as figure 3 
			\State Train with bin$1$
			\State Update $\theta$ with batch loss
			\For {$t= 2 \ to \ T \ldots$}
			\State Compute Nuclear Norm as $\left\{d_t\left(\mathbf{x}^{n}\right)\right\}_{n=1}^{N}$ on current model
			\State Sort $\left\{\mathbf{x}_{t}^{n}\right\}_{n=1}^{N} \ by \left\{d_{t}\left(x^{n}\right)-d_{t-1}\left(x^{n}\right)\right\}_{n=1}^{N}$ from large to small
			\State Divide $\left\{\mathbf{x}_{t}^{n}\right\}_{n=1}^{N}$ into $k$ parts and re-sample as figure 3 
			\State Train with new bins
			\State Update $\theta$ with batch loss
			\EndFor\\
			\Return $\theta$
		\end{algorithmic}
	\end{algorithm}
	
	\subsection{Difficulty Criterion based on model capacity}
	
	Although the above-mentioned difficulty criterion is somewhat reasonable from a linguistic point of view, it may not be the most critical for the model. In other words, the longer the sentence is not necessarily the hardest. The most reliable difficulty criterion should be determined by the model trained under the same task\cite{xu2020curriculum}. Motivated by this, we further introduce a sentence difficulty related to the model capacity based on the initial sentence difficulty. Besides difficulty values are distinguished among different textual samples, we think the difficulty criterion of the same sample is also variable in different learning steps. To this aim, the difficulty criterion $D_{t_i}$ of sample $i$ in $t_{st}$ training step based on model capacity uses the amount of change in the nuclear norm of the relative position sample embedding of the model $E_t$:
	
	\begin{equation}
		d_t^i=\operatorname{tr}\left(\sqrt{{\left(E_t^i\right)}^T {\left(E_t^i\right)}}\right) 
		\label{XX}
	\end{equation}

	\begin{equation}
		D_t^i =  d_t^i-d_{t-1}^i 
		\label{XX1}
	\end{equation}
where $E_{t}$ denotes the last transformer layer of the model at the $t_{st}$  training step, and $i$  denotes the $i_{st}$ training sample after sorting by difficulty score at the $t_{st}$ training step. 

\begin{figure}[ht]
		\centering
		\includegraphics[width=\linewidth]{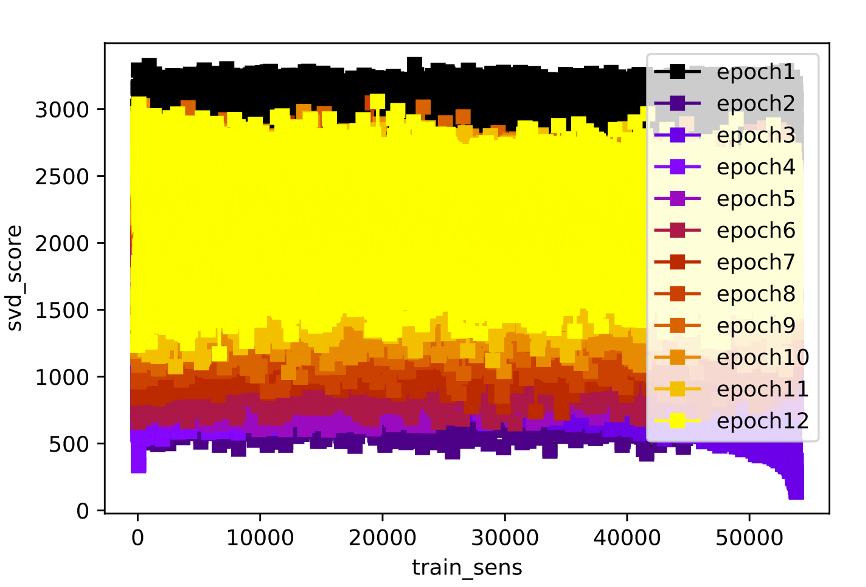}
		\caption{Previous nuclear norm on AAPD train dataset from pretrain to fine-tune process. The trend of the nuclear norm of training samples from pretrained to fine-tuned steps descends firstly along with lowering noise and characters with linear correlation, and then ascends along with learning and extracting features which are more fine-grained.}
		\label{fig:2a}
	\end{figure}

\begin{figure}[ht]
		\centering
		\includegraphics[width=\linewidth]{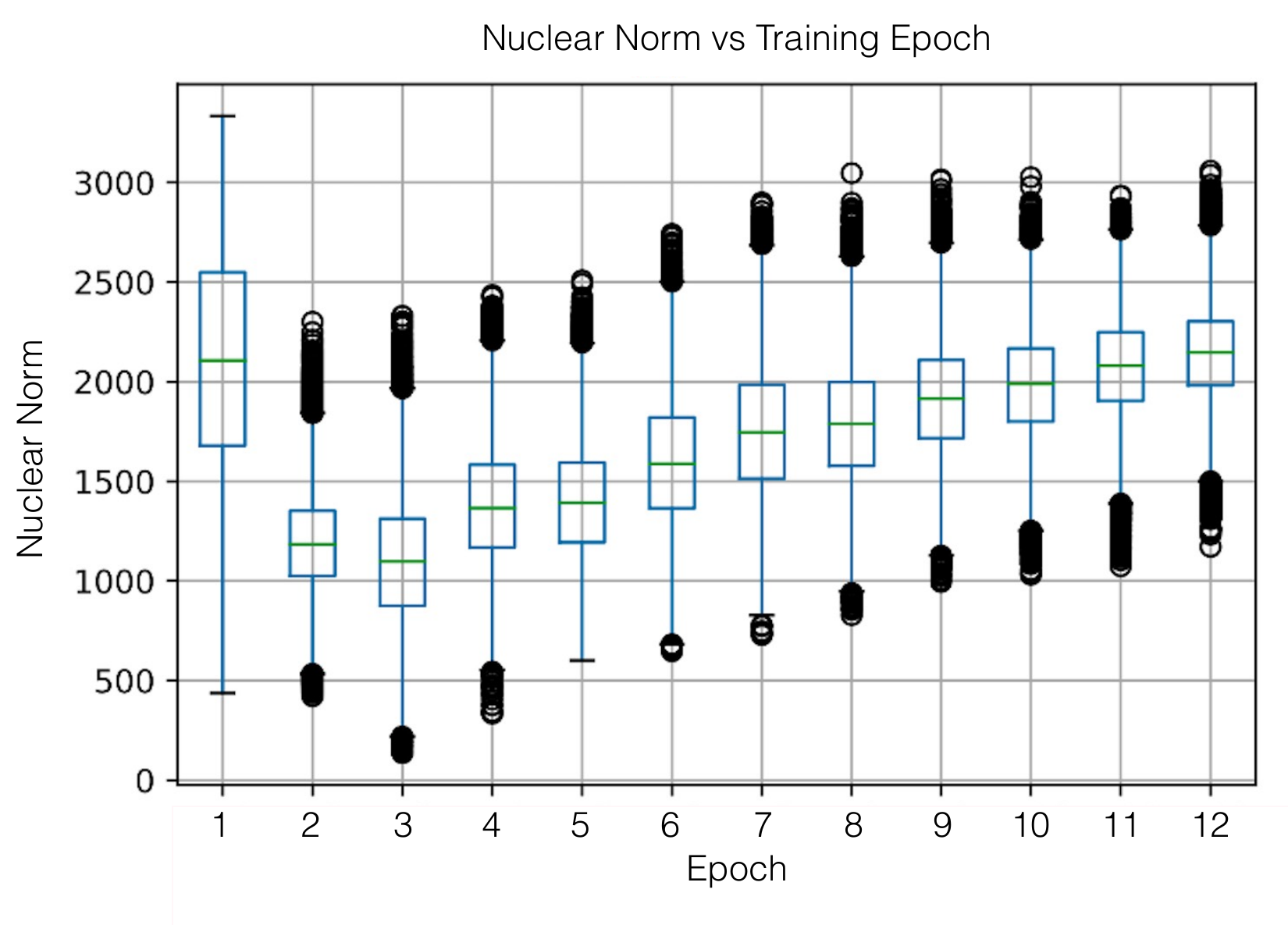}
		\caption{Box-plot of previous nuclear norm on AAPD train dataset from pretrained to fine-tuned process. Compared the current state with the previous state, the less nuclear norm of all relative training samples change, the more difficult the samples will be.}
		\label{fig:2b}
	\end{figure}

\begin{figure}[ht]
		\centering
		\includegraphics[width=\linewidth]{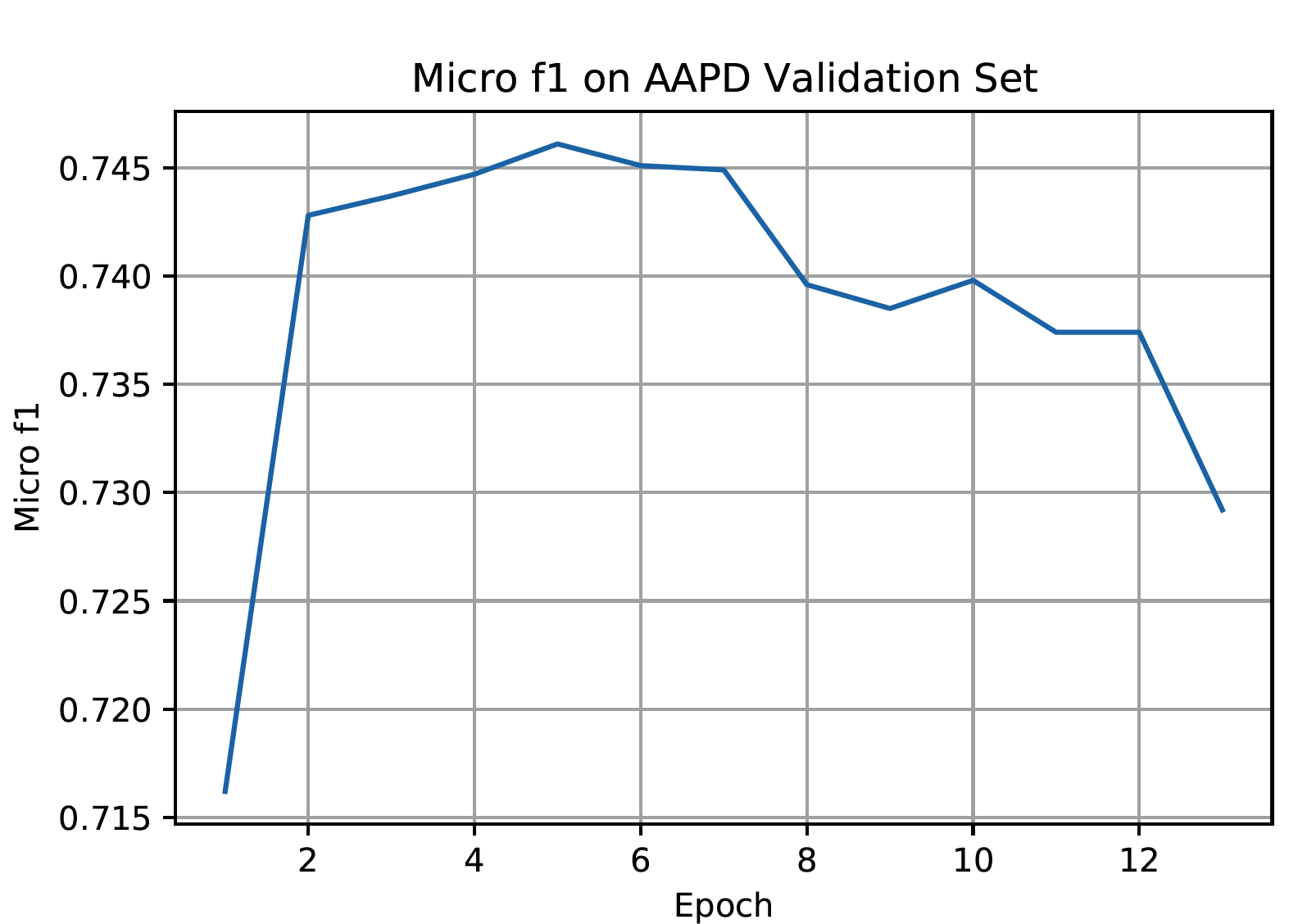}
		\caption{Micro-F1 on AAPD validation dataset. From Figure \ref{fig:2a} and Figure \ref{fig:2b}, there is a close nexus between the trend of the nuclear norm of train data and the precision capacity of the model (micro-f1) on valid data. Compared the current state with the previous state, the less nuclear norm of all relative training samples change, the higher the micro-f1on valid data is, the stronger the model would be. It is clear that the learning ability of the model reaches its highest point at epoch 5.}
		\label{fig:2c}
	\end{figure}

This proposal is motivated by the empirical results shown in Figure \ref{fig:2a},\ref{fig:2b},\ref{fig:2c}, where we show the Micro-F1 scores and the nuclear norms of the sentence embedding matrix at each checkpoint of a model on text classification. Figure \ref{fig:2a} shows that the trend of the nuclear norm of training samples from pretrained BERT to fine-tuning steps descends firstly along with lowering noise and characters with linear correlation, and then ascends along with learning and extracting features which are more fine-grained. And we can find from Figure \ref{fig:2b} and Figure \ref{fig:2c},there is a close nexus between the trend of the nuclear norm $D_{t_{i}}$  of train data and the precision capacity of the model (Micro-F1) on valid data. Compared the current state with the previous state, the less nuclear norm of all relative training samples change, the higher the Micro-F1 on valid data is, the stronger the model would be. It is clear that the learning ability of the model reaches its highest point at epoch 5.

	\subsection{Learning Pace}
	
	 Like in human education, without review after teaching a simple course, students may become confused and forget some of the previous simple knowledge after learning new and harder knowledge. If every year, all the knowledge is given among primary school to high school from easy to difficult, which will consume too much time and might be appropriate for a student. 
	
\begin{figure}[t]
		\centering
		\includegraphics[width=\linewidth]{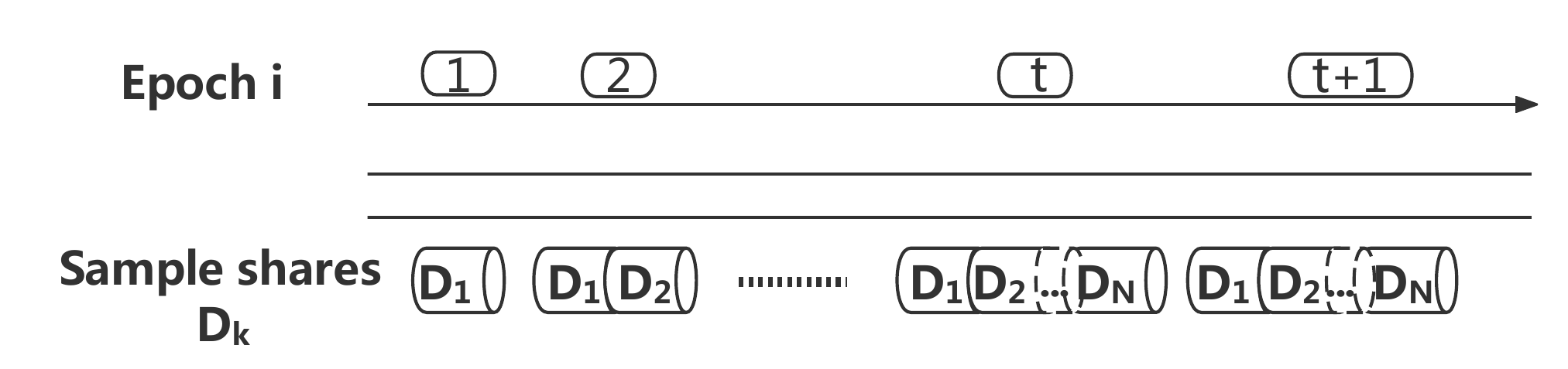}
		\caption{Sampling strategy. In the process of resorting from the easiest to the hardest one and sampling uniformly at each learning stage, we sort the train data D by the difficulty score, and then scatter the train data D after sorting into N shares as at each training epoch.}
		\label{fig:3}
	\end{figure}

	In the process of resorting from the easiest to the hardest one and sampling uniformly at each learning stage, we sort the training dataset $D$ by the difficulty score, and then scatter the trainset $D$ after sorting into N shares as $\left\{\mathbf{D}_{k}\right\}_{k=1}^{N}$ at each training epoch. In this way, we can compare the present easiest sample score with previous easiest sample score and the hardest one now to the hardest one previous. And at training process, we gradually widen the horizons from the easiest to the hardest as below. With the increase in the number of iterations, the data visible to the model is also increasing. After the model can see all the data in some round, it will then use the whole data for training until the model converges, as is shown in Figure \ref{fig:3}
	
	To reduce the impact of noise data and the effects of hyperparameter such as the number of shares, the number of samples in each bucket, and so on, we used all the data inside each share in training to ensure that each phase fully learned about each part. The annealing strategy\cite{xu2020curriculum} is used to continuously review the simple samples that have been learned to enable the model to accept new knowledge while maintaining the ability to master simple knowledge.
	
	\section{Experiments}
	\subsection{Experimental Setting }

	Our experiments are based on Bert4Keras and NVIDIA V100. The version of Bert pretrained model we adapt is English base-uncased. In our training process, we set the same seed as 2, the batch size as 25 and the max length of sequence as 250, the optimizer as Adam \cite{kingma2014adam} with learning rate of 5e-5.

	\subsection{Datasets and Evaluation Metrics}

	We put our experiments on several dataset with different task target. One is large dataset for multi-label classification task called AAPD \cite{yang2018sgm} which is long text about computer science, and the others are small datasets.
	Those small datasets covers multi classification and text matching tasks, that belongs to glue dataset. In AAPD dataset, we divide the data to three parts: train data, test data, and valid data. More detail is in Table \ref{table:datadesp}.While, According to the datasets of glue, we just use train data and valid data that means all the result metrics are based on valid data.

	\begin{table}[H]
		\centering
		\fontsize{8}{13}
		\selectfont
		\caption{Data Description. Y denotes the total number of labels. HL denotes hamming loss. According to the tasks belong to glue, we use F1 and Matthew as the metric for MRPC and CoLA. At the same time, we use Hamming Loss, Micro-F1 and Macro-F1 to measure the result of AAPD.   } 
		\label{table:datadesp}
		\setlength{\tabcolsep}{1.5mm}{
			\begin{tabular}{cccccc}
				\toprule
				Dataset & Y  & Train  & Dev  & Test & Metrics                                                              \\ \midrule
				AAPD    & 54 & 53.84k & 1.0k & 1.0k & \begin{tabular}[c]{@{}c@{}}Accuracy (Acc), \\ Hamming loss (HL), \\ Micro-F1 (Mi-F1), \\ Macro-F1 (Ma-F1)\end{tabular} \\\hline 
				MRPC    & 2  & 3.7k   & 408  &   1.7k   & F1                                                                   \\ 
				CoLA    & 2  & 8.6k   & 1.0k &    1.0k  & Matthew (Mcc)                                                             \\ 
				\bottomrule
		\end{tabular}}
	\end{table}
	
	\subsection{Baseline Methods }
	
	As is shown in the Table \ref{table:datadesp}, according to the tasks belong to glue, we use F1 and Matthew as the metric for MRPC and CoLA. While, for multi-label classification, especially for the long-tail dataset AAPD, we follow previous works \cite{yang2018sgm,Tsai2020OrderfreeLA}that means on the one hand, we use a group of metrics to measure the result of AAPD dataset, such as hamming loss, Micro/Macro-F1 scores, on the other hand, we analyze the Macro-F1 in each label bin that is sorted and organized by label frequency from high to low to judge whether our training strategy is also suitable for low frequency labels. The best weights of model while training is saved by monitoring evaluation metrics as above on the validation set. And the baselines of all the tasks are based on the model without curriculum learning which are implemented on a BERT \cite{devlin2018bert} classifier by us. The result on Bert shows the effectiveness of our SPDCL training strategy.
	
	\subsection{Results and Analysis}

	\begin{table}[t]
		\selectfont
		\caption{Final results on the test set of AAPD, MRPC and CoLA. The model trained with curriculum learning strategy has advantages over the model without curriculum learning} 
		\label{table:results}
		\begin{adjustbox}{width=\columnwidth,center}
		\begin{tabular}{lcccccc}
			\hline
			Dataset      &\multicolumn{4}{c}{AAPD}  & MRPC  & CoLA           \\ \cline{2-5}
			Metric                                               & Mi-F1           & Ma-F1            & Acc            & HL             & F1             & Mcc            \\ \hline
            BERT Base & 0.727          & 0.569          & 0.411          & 0.023          & 0.893          & 0.593          \\ 
			BERT SPDCL(Ours)& \textbf{0.734} & \textbf{0.579} & \textbf{0.415} & \textbf{0.022} & \textbf{0.898} & \textbf{0.610} \\ \hline
		\end{tabular}
		\end{adjustbox}
	\end{table}
	
	We analyze the experimental results of all text classification tasks from different perspectives. In terms of global perspective, the model with curriculum learning strategy while training performs better than the participate without curriculum learning strategy. In terms of local perspective, our method still works in the special extreme imbalanced data, long-tail data, without any model structure changes and data argumentation. The results among the various tasks demonstrates the generalization ability of our method and the effectiveness in imbalanced data.
		\begin{figure}[t]
		\centering
		\includegraphics[width=\linewidth]{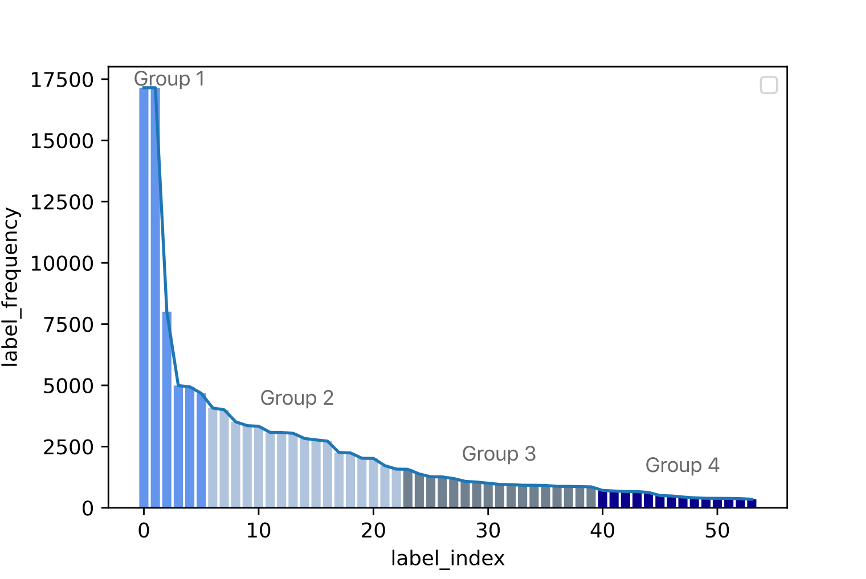}
		\caption{Data distribution on AAPD in terms of label frequency. The long-tail distribution on the AAPD data in terms of label frequency that is sorted and gathered into four groups by their label frequency from high to low.}
		\label{fig:4a}
	\end{figure}
	\FloatBarrier
	\textbf{Main Observation } The final results on the test set (validation set) of the multi-classification task (COLA), texts matching task (MRPC), and multi-label text classification task (AAPD) are shown in Table \ref{table:results}. It can be clearly seen from various indicators that the model with curriculum learning training strategy has advantages over the model without curriculum learning. Specifically, the model based on our training strategy, SPCDL, performs better in all metrics, such as the Matthew increases from 0.593 to 0.609 on the COLA validation, and the F1 increases from 0.893 to 0.898 on the MRPC validation. In terms of AAPD, the hamming loss decreases from 0.023 to 0.022, the Micro-F1 increases from 0.72 to 0.73, Macro-F1 increases from 0.569 to 0.579, and acc increases from 0.411 to 0.415.
	
\begin{figure}[!ht]
		\centering
		\includegraphics[width=\linewidth]{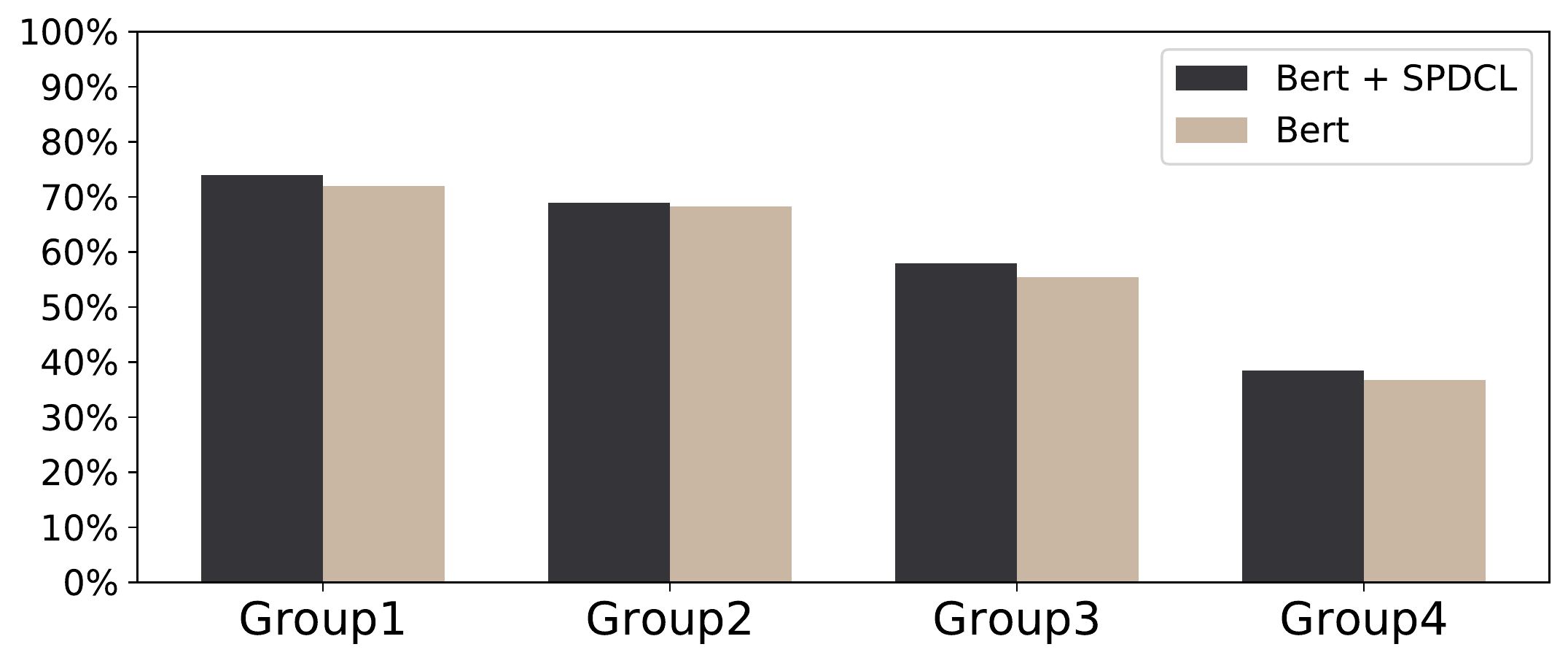}
		\caption{Macro-F1 on different label groups of AAPD test dataset. The SPDCL we proposed matters in imbalanced data, and enhanced all kind of labels, no matter low-frequency labels or high-frequency labels.
		}
		\label{fig:4b}
	\end{figure}

		\textbf{Performance on Long-Tail data} Figure \ref{fig:4a} shows the long-tail distribution on the AAPD data in terms of label frequency that is sorted and gathered into four groups by their label frequency from high to low. From Figure \ref{fig:4b}, we can find that the SPDCL we proposed 
	matters and enhanced all kind of labels, no matter low-frequency labels or high-frequency labels.
	
\section{Conclusions and Future Work  }
	
	We proposed a novel sentence difficulty criterion, consisting of both linguistically motivated features and learning and task-dependent features. Empirical results on the small and large-scale datasets including AAPD, MRPC and CoLA tasks, verify the generalizability and usability of the proposed method, which provides a significant performance boost for text classification tasks in imbalanced settings. 
	We also analyze the performance on long-tail imbalanced datasets that shows the improvement of performance on long-tail data by the training strategy of self-paced dynamic curriculum learning. 
	
	Still there are some potential research directions worth exploring in the future, like optimizing the efficiency of computation, and reducing the computational cost. Another point is to put our training strategy on more downstream tasks, not only text classification. 

	\section{Acknowledgement}
This paper is supported by the Key Research and Development Program of Guangdong Province under grant No.2021B0101400003. Corresponding author is Jianzong Wang from Ping An Technology (Shenzhen) Co., Ltd (jzwang@188.com).

\bibliographystyle{IEEEtran}
\bibliography{mybib}
\end{document}